\documentclass[conference]{IEEEtran}
\IEEEoverridecommandlockouts
\usepackage{cite}
\usepackage{amsmath,amssymb,amsfonts}
\usepackage{algorithmic}
\usepackage[caption=false]{subfig}
\usepackage{graphicx}
\usepackage{textcomp}
\usepackage{xcolor}
\def\BibTeX{{\rm B\kern-.05em{\sc i\kern-.025em b}\kern-.08em
    T\kern-.1667em\lower.7ex\hbox{E}\kern-.125emX}}
\begin{document}

\title{Semi-Supervised Image-to-Image Translation\\}

\author{\IEEEauthorblockN{Manan Oza}
\IEEEauthorblockA{\textit{Department of Computer Engineering} \\
\textit{D. J. Sanghvi College of Engineering}\\
Mumbai, India \\
manan.oza0001@gmail.com}
\and
\IEEEauthorblockN{Himanshu Vaghela}
\IEEEauthorblockA{\textit{Department of Computer Engineering} \\
\textit{D. J. Sanghvi College of Engineering}\\
Mumbai, India \\
himanshuvaghela1988@gmail.com}
\and
\IEEEauthorblockN{Prof. Sudhir Bagul}
\IEEEauthorblockA{\textit{Department of Computer Engineering} \\
\textit{D. J. Sanghvi College of Engineering}\\
Mumbai, India \\
Sudhir.Bagul@djsce.ac.in}
}

\maketitle

\begin{abstract}
Image-to-image translation is a long-established and a difficult problem in computer vision. In this paper we propose an adversarial based model for image-to-image translation. The regular deep neural-network based methods perform the task of image-to-image translation by comparing gram matrices and using image segmentation which requires human intervention. Our generative adversarial network based model works on a conditional probability approach. This approach makes the image translation independent of any local, global and content or style features. In our approach we use a bidirectional reconstruction model appended with the affine transform factor that helps in conserving the content and photorealism as compared to other models. The advantage of using such an approach is that the image-to-image translation is semi-supervised, independant of image segmentation and inherits the properties of generative adversarial networks tending to produce realistic. This method has proven to produce better results than Multimodal Unsupervised Image-to-image translation.
\end{abstract}


\section{Introduction}
Image-to-image transfer has established itself as an important domain in computer vision since the first paper published by Gatys et al. \cite{b1}. Also known as Neural Style Transfer, it has had many variations over the years, image colorization \cite{b14}, style transfer \cite{b1}, image-to-image transfer \cite{b3} and so on. For which generally deep neural networks have been used with architectural variances. For instance, we can make a day time image (also known as the content image) of a city look like a night time image by selecting the appropriate style (reference) image.  Likewise we can have diverse types of features transfered from one image to another which include time, color, seasonal translations as well.

Image-to-image translation is the process of translating one image onto another while preserving the content and photorealism of the original content image. Deep-learning techniques have proved excellent in faithful and photorealistic style translation \cite{b1, b2, b3, b7}. Our approach is built upon the idea of generative adversarial networks introduced by Goodfelow et al. \cite{b8}. The underlying concept of such a neural network architecture is that a GAN consists of a generator and a discriminator. The discriminator is trained to identify real images while the generator tries to fool the discriminator by creating counterfiet images from noise and passes them on to the discriminator. Which then returns a verdict on how close the counterfiet images are to a real one. Based on this feedback the generator improves itself and creates another image and the cycle repeats.

Here in our paper we make use of an improvised GAN architecture appended with an Affine Loss factor calculated from a Matting Laplacian matrix \cite{b6} in the final loss function. This additionl factor helps in maintaining spatial integrity and preserve photorealism in the content image. Since generative adversarial networks create images from noise they are prone to distortions and noisy images but provide with the biggest advantage, they do not form the basis of simple color and style mapping. They recreate the content image with the style variations.

\section{Related Work}
Image-to-image style transfer has reached state-of-the-art \cite{b2, b3, b7} results. The current existing algorithms work in either of the two broadly divided classes: local translation and global translation. But neither of the algorithms excel in both photorealism and faithful style translation at the same time and for all test cases. One or the other factor gets compromised. Global stylization methods work by matching statistical factors of the pixel values \cite{b11} whereas local stylization is achieved by algorithms that find close and consistent relations between pixel values of the content and style images. Another classification is based on the algorithm's ability to translate low-level and (or) high-level features. Low-level features translation involves preservation of the intricacies in the content image while modifying the color or position with respect to the style image. Whereas high-level feature translation is the mapping of broader features which by example means day to night, summer to winter translations.

The best works proposed by Luan et al. \cite{b2} and Li et al. \cite{b7} are based on the paradigm of matching the gram matrices and makes use of semantic segmentaions of the content and style images. Which take in only the content and style images as the inputs for the network. These algorithms perform post-processing like affine smoothing techniques thereby drastically improving the quality of the resultant images. Such methodologies make use of segmented images derived from the content and style images and then perform style translations from one segment to another by comparing the gram matrices of the input images. Other such algorithms based on a similar paradigm are proposed by Gatys et al. \cite{b1}, Huang et al. \cite{b2} and many others, \cite{b1, b7, b12}. 

Promising results have been showcased by various GAN architectures namely Pix2pix by Isola et al. \cite{b13}, Unsupervised Image-to-image Translation by Liu et al. \cite{b4}, CycleGAN and BicycleGAn by Zhu et al. \cite{b5}. All of which take in a dataset consisting of multiple images similar to the content and the style domain. Multimodal Unsupervised Image-to-Image Translation by Huang et al. \cite{b3} provides an approach to the problem by narrowing down the content domain to only one image and a number of style images which constitute the style latent code \cite{b3}. They have proposed that to make the translation unsupervised the syle images are decomposd into a common style latent space. The content space is sampled from this style space based on a conditional distribution to perfom the translation.

In our proposal we narrow down our method to one content and one style image which does not make it completely unsupervised as there is only one target style image. We use the same architecture as proposed by Huang et al. \cite{b3} with an additional affine loss factor added to the loss function which adds to the smoothness and faithful style transfer which are combined with the properties of generative adversarial networks.

\section{Methodology}
In addition to the model proposed by Huang et al. \cite{b3} we add the local affine transfrom $\mathcal{L}_m$ also known as the photorealism factor of the content image calculated from the Matting Laplacian matrix proposed by Levin et al. \cite{b6}.

\subsection{Assumptions}
All assumptions are exactly the same as that made in the paper Multimodal Unsupervised Image-to-Image Translation by Huang et al. \cite{b3} which are as follows.
The model assumes that the content and style images are composed of distinct image spaces \textit{$x_i \in$ $\mathcal{X}_i$} where \textit{$x_i$} is the \textit{$i^{th}$} image and \textit{$\mathcal{X}_i$} is its corresponding image space. Here our goal is to estimate the conditional distributions \textit{$p(x_1 \vert x_2)$} and \textit{$p(x_2 \vert x_1)$} leading to the learned translation models \textit{$p(x_{1\longrightarrow2} \vert x_2)$} and \textit{$p(x_{2\longrightarrow1} \vert x_1)$} respectively given that \textit{$p(x_1)$} and \textit{$p(x_2)$} are the marginal distributions of \textit{$x_1$} and \textit{$x_2$} respectively.

We make another assumption that \textit{$x_i \in$ $\mathcal{X}_i$} is composed of a content latent space \textit{$c \in$ $\mathcal{C}$} and a style latent space \textit{$s_i \in$ $\mathcal{S}_i$} corresponding to every image from the dataset. Thus two images \textit{$(x_1, x_2)$} are generated from the individual generators by \textit{$x_1 =  G_1^*(c,s_2)$} and \textit{$x_2 =  G_2^*(c,s_1)$}. \textit{$G_1^*$} and \textit{$G_2^*$} are generator functions with \textit{$E_1^*$} and \textit{$E_2^*$} being their inverse encoders where \textit{$E_1^* = (G_1^*)^{-1}$} and \textit{$E_2^* = (G_2^*)^{-1}$}. Hence our aim is to train the encoder and generator functions using neural networks.

\subsection{Matting Laplacian}
Image matting is the process of extracting the foreground and the background from an image with minimal possible user intervention. The Matting Laplacian \cite{b6} process produces an alpha matte which is the segmented image with the foreground object in white and the background in black or vice versa as per the requirements. Using this matting laplacian matrix we calculate the local affine transform factor $\mathcal{L}_m$ also known as the photorealism factor.
\begin{equation}
\begin{aligned}
\mathcal{L}_m = \sum_{c=1}^{3} V_c[O]^T\mathcal{M}_IV_c[O]
\end{aligned}
\end{equation}
It is a summation of the affine losses of all the three channels of the image. $\mathcal{M}_I$ is the least-squares penalty function that is dependant on the input image \textit{I}. The dimensions of the $\mathcal{M}_I$  matrix are (\textit{N $\times$ N}) and $V_c[O]$ is the vectorized format of the input image \textit{O} in the channel \textit{c} having dimensions (\textit{N $\times$ 1}). Thus this factor proves crucial in preserving the photorealism and the content image in our proposal.

\begin{figure}[htp]
\subfloat[Single domain reconstruction of \textit{$x_1$}]{
  \includegraphics[width=0.5\textwidth]{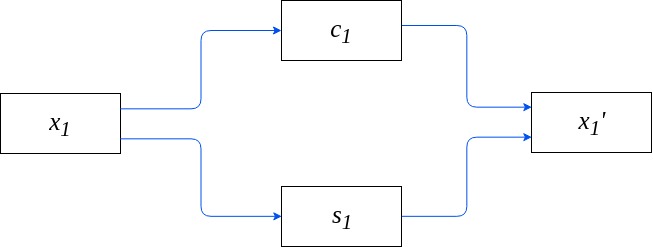}
}\\
\subfloat[Single domain reconstruction of \textit{$x_2$}]{
  \includegraphics[width=0.5\textwidth]{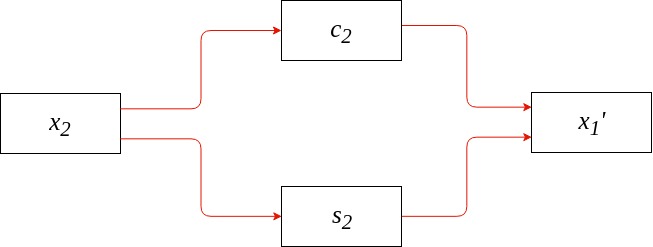}
}
\caption{The above given images are the representations of how the self domain reconstruction of our model works. The images \textit{$x_1$} and \textit{$x_2$} are encoded into their respective content and style latent codes \textit{$c_i$} and \textit{$s_i$}. The reconstructed images \textit{$x_1{'}$} and \textit{$x_2{'}$} are not equal to their corresponding input imags because of $\mathcal{L}_1$ loss.} 
\end{figure}

\begin{figure}[htp]
\includegraphics[width=0.5\textwidth]{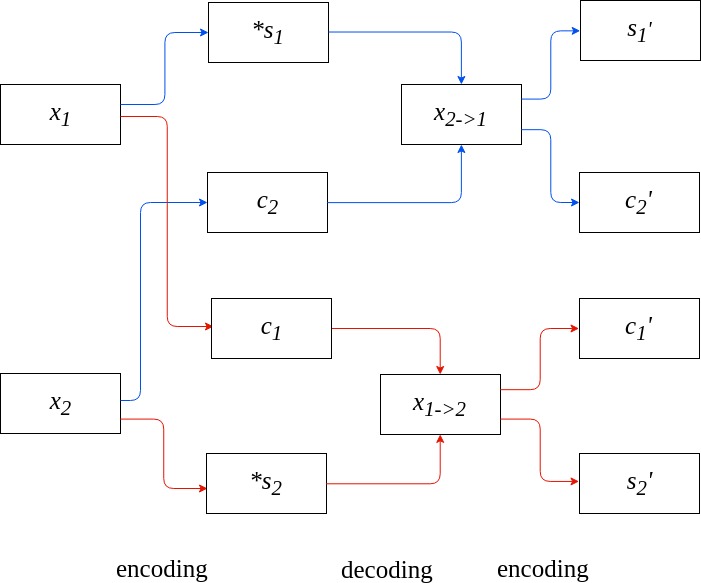}
\caption{This image represents the cross-domain translation of \textit{$x_{1\longrightarrow2}$} and \textit{$x_{2\longrightarrow1}$}. The '*' represents the Gaussian prior. We encounter $\mathcal{L}_1$ losses when reconstructing images from \textit{$s_i$} to \textit{$s_i{'}$} and \textit{$c_i$} to \textit{$c_i{'}$} thereby fulfilling the bidirectional reconstruction properties of our model. GAN loss is encountered when the translation of \textit{$x_1$} $\rightarrow$ \textit{$x_{2\longrightarrow1}$} and \textit{$x_2$} $\rightarrow$ \textit{$x_{1\longrightarrow2}$} takes place. }
\end{figure}

\begin{figure*}
 \center
  \includegraphics[height=200pt, width=\textwidth]{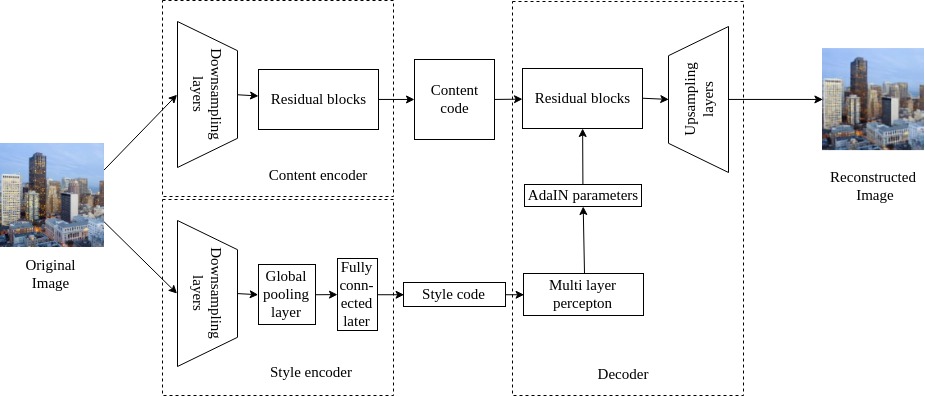}
  \caption{The auto-encoder architecture. It consists of a content encoder, style encoder and a decoder. The content encoder comprises of three convolutional blocks which perform downsampling, followed by four residual blocks. The style encoder comprises of five convolutional layers followed by a global average pooling layer followed by a fully connected layer at the end. The decoder makes use of a multi-layer perceptron that generates AdaIN \cite{b15} parameters from the style code. The content code along with AdaIN parameters is processed by four residual blocks. The output of the residual blocks is passed on to three upsampling layers that generate the final image.}
  \label{aen}
\end{figure*}

\subsection{Model}
Our model given in figure \ref{aen} constitutes an encoder and a decoder \textit{$E_i^*$} and \textit{$G_i^*$} respctively for every domain $\mathcal{X}_i$, in our case \textit{$i$} = 1, 2. The encoder is factorized from the content and style latent codes \textit{$c_i$} and \textit{$s_i$}.
\begin{equation}
\begin{aligned}
(\textit{$c_i$},\textit{$s_i$}) = (E_i^c(\textit{$x_i$}),E_i^s(\textit{$x_i$})) = E_i(\textit{$x_i$})
\end{aligned}
\end{equation}

Thus for image-to-image translation we interchange the encoders and decoders i.e. for translation \textit{$x_{1\longrightarrow2}$} we make use of the content code \textit{$c_1$} = \textit{$E_1^c(x_1)$} and a randomly drawn style latent code from \textit{$s_2$}. Subsequently we use the decoder \textit{$G_2$} to generate the image.
\begin{equation}
\begin{aligned}
\textit{$x_{1\longrightarrow2}$} = \textit{$G_2(c_1, s_2)$}
\end{aligned}
\end{equation}

The loss function is composed of two factors, the bidirectional reconstruction loss and the adversarial loss. The bidirectional reconstruction loss is added to make sure that there is a two way reconstruction of images in the directions, image $\rightarrow$ latent $\rightarrow$ image and latent $\rightarrow$ image $\rightarrow$ latent. The image reconstruction loss is computed as the difference between the image reconstructed from the latent spaces \textit{$c_1$} and \textit{$s_1$} of image \textit{$x_1$} and the image \textit{$x_1$} which is given by (it is similar to $\mathcal{L}_{recon}^{x_2}$ for the image \textit{$x_2$}):
\begin{equation}
\begin{aligned}
\mathcal{L}_{recon}^{x_1} = \mathbb{E}_{x_1 \sim p(x_1)}[\Vert G_1(E_1(x_1)) - x_1\Vert_1]
\end{aligned}
\end{equation}
The latent reconstruction loss $\mathcal{L}_{recon}^{c_1}$ is the difference between the content encoding of the generated image \textit{$G_2(c_1, s_2)$} and the content encoding \textit{$c_1$} of the image \textit{$x_1$} and $\mathcal{L}_{recon}^{s_2}$ is the difference between the style encoding of the generated image \textit{$G_2(c_1, s_2)$} and the style encoding \textit{$s_2$} of the image \textbf{$x_2$} they are given by the equations (which are similar for and $\mathcal{L}_{recon}^{s_1}$):
\begin{equation}
\begin{aligned}
\mathcal{L}_{recon}^{c_1} = \mathbb{E}_{c_1 \sim p(c_1), s_2 \sim p(s_2)}[\Vert E_2^c(G_2(c_1,s_2)) - c_1\Vert_1]
\end{aligned}
\end{equation}
\begin{equation}
\begin{aligned}
\mathcal{L}_{recon}^{s_2} = \mathbb{E}_{c_1 \sim p(c_1), s_2 \sim p(s_2)}[\Vert E_2^s(G_2(c_1,s_2)) - s_2\Vert_1]
\end{aligned}
\end{equation}
Here \textit{$q(s_2)$} is defined as the prior $\mathcal{N}$(0, \textbf{I}) and \textit{$p(c_1)$} is defined as \textit{$c_1 = E_1^c(x_1)$} where \textit{$x_1 \sim p(x_1)$}.

Since we use a GAN framework we encounter an adversarial loss which is supposed to be minimised so that the generated images are as identical as possible to the original images. This loss is given by:
\begin{equation}
\begin{aligned}
\mathcal{L}_{GAN}^{x_2} = \mathbb{E}_{c_1 \sim p(c_1), s_2 \sim p(s_2)}[log(1-D_2(G_2(c_1,s_2)))] + \\
\mathbb{E}_{x_2 \sim p(x_2)}[log D_2(x_2)]
\end{aligned}
\end{equation}
Here \textit{$D_2$} is the discriminator function that distinguishes between the real image \textit{$x_2$} and the translated images. The discriminator function \textit{$D_1$} and loss $\mathcal{L}_{GAN}^{x_2}$ are defined in a similar way.

As mentioned earlier the architecture we use is essentially the same as that was proposed by Huang et al. \cite{b3}. The only difference being that in our approach we use only one style image and add the affine transform loss in the overall loss function. We assume \cite{b2} that the input images are photorealistic and we do not have to lose this property. Thus we penalize the loss fuction with the photorealism factor so as not to lose this property while minimizing the reconstruction losses from the image, content and style latent spaces. The overall loss fuction proposed by us is given by:
\begin{equation}
\begin{aligned}
\textstyle\underset{E_1,E_2,G_1,G_2}{\min_{\vphantom{p}}} \underset{D_1,D_2}{\max_{\vphantom{p}}} \mathcal{L}(E_1,E_2,G_1,G_2,D_1,D_2) = \\ \mathcal{L}_{GAN}^{x_1} + \mathcal{L}_{GAN}^{x_2} + \lambda_x(\mathcal{L}_{recon}^{x_1} + \mathcal{L}_{recon}^{x_2}) + \\ \lambda_c(\mathcal{L}_{recon}^{c_1} + 
\mathcal{L}_{recon}^{c_2}) + \lambda_s(\mathcal{L}_{recon}^{s_1} + \mathcal{L}_{recon}^{s_2}) + \\ \lambda_A(\mathcal{L}_m^{x_1} + \mathcal{L}_m^{x_2})
\end{aligned}
\end{equation}
Where $\lambda_x$, $\lambda_c$, $\lambda_s$ are the weights that control the reconstruction, and $\lambda_A$ is the photorealism regularization weight \cite{b2}.

\begin{figure*}[htp]
\subfloat[]{
  \includegraphics[width=0.15\textwidth]{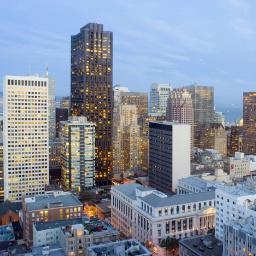}
  \label{a}
}
\subfloat[]{
  \includegraphics[width=0.15\textwidth]{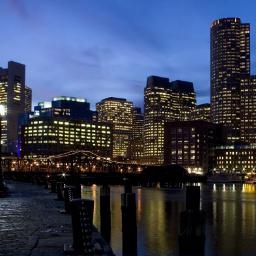}
  \label{b}
}
\subfloat[]{
  \includegraphics[width=0.15\textwidth]{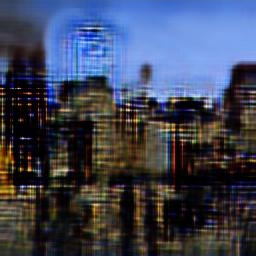}
  \label{c}
}
\subfloat[]{
  \includegraphics[width=0.15\textwidth]{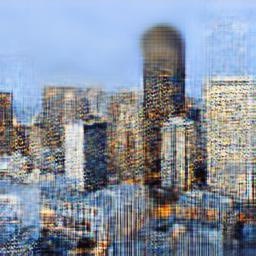}
  \label{d}
}
\subfloat[]{
  \includegraphics[width=0.15\textwidth]{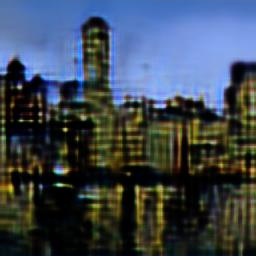}
  \label{e}
}
\subfloat[]{
  \includegraphics[width=0.15\textwidth]{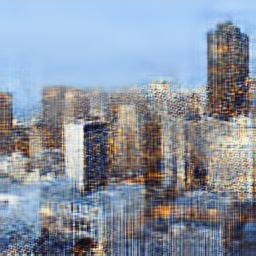}
  \label{f}
}\\
\subfloat[]{
  \includegraphics[width=0.15\textwidth]{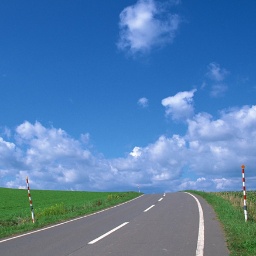}
  \label{g}
}
\subfloat[]{
  \includegraphics[width=0.15\textwidth]{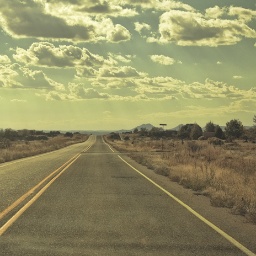}
  \label{h}
}
\subfloat[]{
  \includegraphics[width=0.15\textwidth]{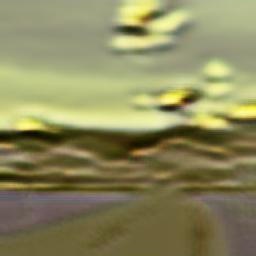}
  \label{i}
}
\subfloat[]{
  \includegraphics[width=0.15\textwidth]{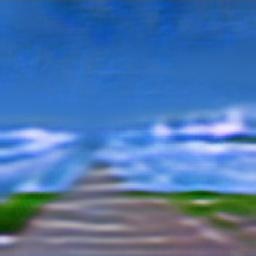}
  \label{j}
}
\subfloat[]{
  \includegraphics[width=0.15\textwidth]{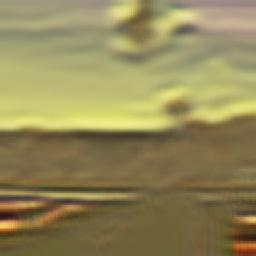}
  \label{k}
}
\subfloat[]{
  \includegraphics[width=0.15\textwidth]{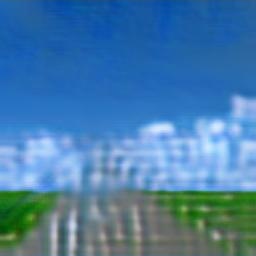}
  \label{l}
}\\
\subfloat[]{
  \includegraphics[width=0.15\textwidth]{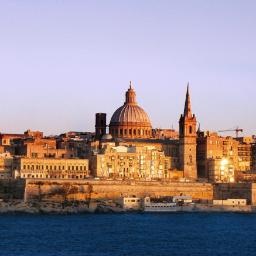}
  \label{m}
}
\subfloat[]{
  \includegraphics[width=0.15\textwidth]{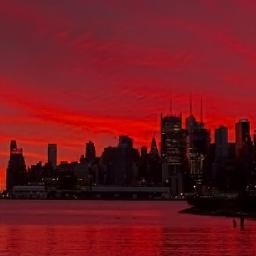}
  \label{n}
}
\subfloat[]{
  \includegraphics[width=0.15\textwidth]{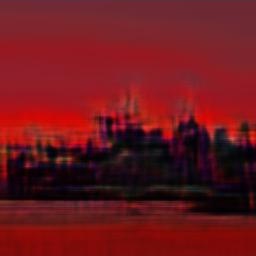}
  \label{o}
}
\subfloat[]{
  \includegraphics[width=0.15\textwidth]{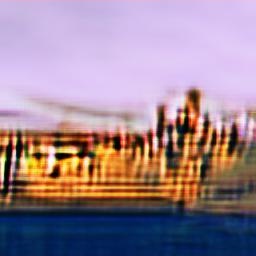}
  \label{p}
}
\subfloat[]{
  \includegraphics[width=0.15\textwidth]{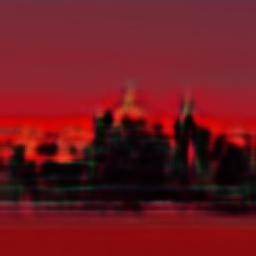}
  \label{q}
}
\subfloat[]{
  \includegraphics[width=0.15\textwidth]{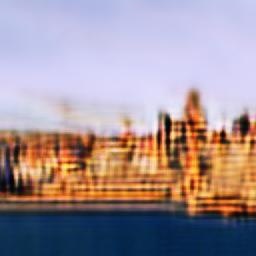}
  \label{r}
}
\caption{First two images in each row i.e. [\ref{a} and \ref{b}], [\ref{g} and \ref{h}] and [\ref{m} and \ref{n}] are the constituent datasets. Image \ref{c} and \ref{i} and \ref{o} are the results obtained from Multimodal Unsupervised Image-to-image Translation \cite{b3} with \ref{a}, \ref{g} and \ref{m} as the content images respectively and \ref{b}, \ref{h} and \ref{n} as the style images respectively. While \ref{d}, \ref{j} and \ref{p} are the results produced by the same method with \ref{b}, \ref{h} and \ref{n} as the content images respectively and \ref{a}, \ref{g} and \ref{m} as the style images respectively. Image \ref{e} is our result with \ref{a} as the content image and \ref{b} as the style image. Image \ref{f} is our result with \ref{b} as the content image and \ref{a} as the style image. Image \ref{k} is our result with \ref{g} as the content image and \ref{h} as the style image. Image \ref{l} is our result with \ref{h} as the content image and \ref{g} as the style image. Image \ref{q} is our result with \ref{m} as the content image and \ref{n} as the style image. Image \ref{r} is our result with \ref{n} as the content image and \ref{m} as the style image.}
\label{results}
\end{figure*}

\subsection{Analysis}
Our goal is to minimize the loss function defined in equation (8). This minima is the optimal state of of our model and at this point the following states are achieved:
\begin{equation}
\begin{aligned}
p(c_1) = p(c_2)
\end{aligned}
\end{equation}
\begin{equation}
\begin{aligned}
p(s_1) = q(s_1)
\end{aligned}
\end{equation}
\begin{equation}
\begin{aligned}
p(s_2) = q(s_2)
\end{aligned}
\end{equation}
\begin{equation}
\begin{aligned}
p(x_1, x_{1\longrightarrow2}) + p(c_1) = p(x_{2\longrightarrow1}, x_2) + p(c_2)
\end{aligned}
\label{eq12}
\end{equation}
The equation \eqref{eq12} is different from the one proposed by Huang et al. \cite{b3} because our model adds the local affine loss of the content images. Our model is constructed in such a way that when \textit{$x_1$} is the content image \textit{$x_2$} is taken as the style image and vice versa. Which is why the local affine loss of both the images is taken into consideration in equation (8) and also the content marginal distributions are added and taken into account when comparing the joint distributions \textit{$p(x_1, x_{1\longrightarrow2})$} and \textit{$p(x_{2\longrightarrow1}, x_2)1$}. At this state the content marginal distributions \textit{$p(c_1)$} and \textit{$p(c_2)$} also become equal. Also at this optimal state the style marginal distributions \textit{$p(s_i)$} are equal to their prior distributions \textit{$q(s_i)$}. The fact that we use one one-to-one image mapping makes our process sound like it follows the supervised learning paradigm, but it does not. Even though we have only one image in the content and style domain the images are encoded into a content and style latent space and translated on the basis of conditional probability. Thus our method is free from any deterministic translations as performed by the methods \cite{b1, b2, b7, b12, b14} which make use of image segmentation that helps in mapping regions of interest in both the content and style images.

\section{Implementation Details}
We have adapted the publicly available pytorch implementation of Multimodal Unsupervised Image-to-image Translation \cite{b3}. The architecture consists of an auto-encoder (generator) and a discriminator. The auto-encoder comprises of a separate content and style encoder and a combined decoder. The auto-encoder architecture consists of the following layers:
\begin{itemize}
\item The content encoder whose content makes up the content latent space (in the listed order):
\begin{itemize}
\item 7 $\times$ 7 convolutional block with stride 1 and 64 filters.
\item 4 $\times$ 4 convolutional block with stride 2 and 128 filters.
\item 4 $\times$ 4 convolutional block with stride 2 and 256 filters.
\item 4 residual blocks each consisting of two 3 $\times$ 3 convolutional blocks with 256 filters.
\end{itemize}
\item The style encoder whose output is added to the style latent space (in the listed order):
\begin{itemize}
\item 7 $\times$ 7 convolutional block with stride 1 and 64 filters.
\item 4 $\times$ 4 convolutional block with stride 2 and 128 filters.
\item 3 4 $\times$ 4 convolutional block with stride 2 and 256 filters.
\item Global average pooling layer.
\item Fully connected layer with 8 filters.
\end{itemize}
\item The decoder which reconstructs an image from the content and style latent code (in the listed order):
\begin{itemize}
\item 4 residual blocks each consisting of two 3 $\times$ 3 convolutional blocks with 256 filters.
\item 2 $\times$ 2 nearest-neighbour upsampling layer followed by a 5 $\times$ 5 convolutional layer with stride 1 and 128 filters.
\item 2 $\times$ 2 nearest-neighbour upsampling layer followed by a 5 $\times$ 5 convolutional layer with stride 1 and 64 filters.
\item 7 $\times$ 7 convolutional block with stride 1 and 3 filters.
\end{itemize}
\end{itemize}
The discriminator used is a multi-scale discriminator proposed by Wang et al. \cite{b9} which makes use of the LSGAN objective function proposed by Mao et al. \cite{b10}. This helps to pilot the generator towards producing realistic and perfom effective translation while preserving the content. The architecture consists of the following layers in the listed order:
\begin{itemize}
\item 4 $\times$ 4 convolutional block with stride 2 and 64 filters.
\item 4 $\times$ 4 convolutional block with stride 2 and 128 filters.
\item 4 $\times$ 4 convolutional block with stride 2 and 256 filters.
\item 4 $\times$ 4 convolutional block with stride 2 and 512 filters.
\end{itemize}

We use the python implementation to compute the Matting Laplacian matrix \cite{b16} from the tensorflow implementation of Deep Photo Style Transfer \cite{b2}. The image, content and style reconstruction weights and the photorealism regularization weight are experimentally set to $\lambda_x$ = 10, $\lambda_c$ = 1, $\lambda_s$ = 1 and $\lambda_A$ = $10^4$ \cite{b2} respectively. Our implementation is available on \textit{$https://github.com/ozamanan/semisit$}.

\section{Results}
Our dataset is composed of only two 3 channel images with resolution 256 $\times$ 256. Thus we use a batch size of 1. Furthermore for every iteration both the images from the dataset are used once to train the respective parts of the network. At once when one image is used as the content image the other one is used as the style image and vice versa thereby completing the bidirectional reconstruction process. All images used for experimental purposes are taken from the implementation of Deep Photo Style Transfer \cite{b2}.

The images fig. \ref{e}, \ref{f}, \ref{k}, \ref{l}, \ref{q} and \ref{r} shown in fig. \ref{results} are the results generated from our code whereas the images \ref{c}, \ref{d}, \ref{i}, \ref{j}, \ref{o} and \ref{p} are the results generated using the code of Huang et al. \cite{b3}. The results shown by us are the optimal results beyond which the images tend to converge to their respective style images. The optimal solution is the one where the resultant image holds the properties of both the content and style images while still being recognised by the discriminator as a constituent image of the dataset. This optimal state is mentioned in eq. \eqref{eq12}.

This optimal state clearly shows an improvement in content preservation and image smoothness over the proposal of Huang et al. \cite{b3}. It is achieved due to the addition of the affine transform factors $\mathcal{L}_m^{x_1}$ and $\mathcal{L}_m^{x_2}$. Thus our proposed methodology generates results that are better in comparison to the results from the method used by Huang et al. \cite{b3}.

\section{Conclusion}
We have proposed an architecture that performs the task of unsupervised image-to-image translation with better accuracy and results. The future work includes reducing the noise and making the results more accurate even for low resolutions. Another future scope lies in broadening this architecture for the generation of music, text and videos.


@inproceedings{huang2018munit,
  title={Multimodal Unsupervised Image-to-image Translation},
  author={Huang, Xun and Liu, Ming-Yu and Belongie, Serge and Kautz, Jan},
  booktitle={ECCV},
  year={2018}
}


\begin{thebibliography}{00}
\bibitem{b1} L. A. Gatys, A. S. Ecker, and M. Bethge. Image style transfer using convolutional neural networks. In Proceedings of the IEEE Conference on Computer Vision and Pattern Recognition, pages 2414–2423, 2016.

\bibitem{b2} Luan, F., Paris, S., Shechtman, E., Bala, K.: Deep photo style transfer. In: CVPR. (2017) 

\bibitem{b3} Huang, X., Liu, M.Y., Belongie, S., Kautz, J.: Multimodal unsupervised image-toimage translation. In: ECCV. (2018)

\bibitem{b4} Liu, M.Y., Breuel, T., Kautz, J.: Unsupervised image-to-image translation networks. In: NIPS. (2017)

\bibitem{b5} Almahairi, A., Rajeswar, S., Sordoni, A., Bachman, P., Courville, A.: Augmented cyclegan: Learning many-to-many mappings from unpaired data. arXiv preprint arXiv:1802.10151 (2018)

\bibitem{b6} A. Levin, D. Lischinski, and Y. Weiss. A closed-form solution to natural image matting. IEEE Transactions on Pattern Analysis and Machine Intelligence, 30(2):228–242, 2008.

\bibitem{b7} Y. Li, M.-Y. Liu, X. Li, M.-H. Yang, and J. Kautz, “A closed-form solution to photorealistic image stylization,”
arXiv:1802.06474, 2018.

\bibitem{b8} Goodfellow, I., Pouget-Abadie, J., Mirza, M., Xu, B., Warde-Farley, D., Ozair, S., Courville, A., Bengio, Y.: Generative adversarial nets. In: NIPS. (2014)

\bibitem{b9} Wang, T.C., Liu, M.Y., Zhu, J.Y., Tao, A., Kautz, J., Catanzaro, B.: High resolution image synthesis and semantic manipulation with conditional gans. In: CVPR. (2018)

\bibitem{b10} Mao, X., Li, Q., Xie, H., Lau, Y.R., Wang, Z., Smolley, S.P.: Least squares generative adversarial networks. In: ICCV. (2017)

\bibitem{b11} Erik Reinhard, Michael Ashikhmin, Bruce Gooch, and Peter Shirley. Color transfer between images. IEEE Computer Graphics and Applications, 21(5):34–41, 2001.

\bibitem{b12} Ulyanov, V. Lebedev, A. Vedaldi, and V. Lempitsky. Texture networks: Feedforward synthesis of textures and stylized images. In International Conference on Machine Learning (ICML), 2016.

\bibitem{b13} Phillip Isola, Jun-Yan Zhu, Tinghui Zhou, and Alexei A Efros. Image-to-image translation with conditional adversarial networks. 2016.

\bibitem{b14} Paulina Hensman and Kiyoharu Aizawa. cgan-based manga colorization using a single training image. arXiv:1706.06918, 2017.

\bibitem{b15} Huang, X., Belongie, S.: Arbitrary style transfer in real-time with adaptive instance normalization. In: ICCV. (2017)

\bibitem{b16} Martin Benson. [Online]. Available: \textit{$https://github.com/$\\$martinbenson/deep-photo-styletransfer/blob/master/$\\$deep\_photo.py$}

\end{thebibliography}
\end{document}